\documentclass[runningheads]{llncs}
\usepackage{graphicx}

\usepackage{tikz}
\usepackage{comment}
\usepackage{amsmath} 
\usepackage{amssymb} 
\usepackage{color}

\usepackage[accsupp]{axessibility}  

\usepackage{amsmath,graphicx}
\usepackage{booktabs}
\usepackage{nicematrix}

\usepackage{multirow}
\usepackage{tabto}
\usepackage{float}
\usepackage[colorlinks,
            linkcolor=blue,       
            anchorcolor=blue,  
            citecolor=blue,    
            ]{hyperref}
\newcommand {\R}{\mathbb{R}}
\usepackage{caption} 
\usepackage{array}
\usepackage{xcolor}
\usepackage{color, colortbl}
	
\definecolor{Gray}{gray}{0.9}
\definecolor{LightCyan}{rgb}{0.88,0.95,1}

\makeatletter
\newcommand*{\@rowstyle}{}
\newcommand*{\rowstyle}[1]{
  \gdef\@rowstyle{#1}%
  \@rowstyle\ignorespaces%
}
\newcolumntype{=}{
  >{\gdef\@rowstyle{}}%
}
\newcolumntype{+}{
  >{\@rowstyle}%
}
\makeatother

\usepackage{amssymb}
\usepackage{pifont}
\newcommand{\xmark}{\ding{55}}%

\begin{document}
\pagestyle{headings}
\mainmatter
\def\ECCVSubNumber{2605}  

\title{
GRIT: Faster and Better Image captioning Transformer Using Dual Visual Features}

\titlerunning{GRIT: Grid- and Region-based Image captioning Transformer}
%
\author{Van-Quang Nguyen\inst{1}
\and
Masanori Suganuma\inst{2,1}
\and
Takayuki Okatani\inst{1,2}
}
\authorrunning{Van-Quang Nguyen et al.}
%
\institute{Graduate School of Information Sciences, Tohoku University\and
RIKEN Center for AIP \\
\email{\{quang,suganuma,okatani\}@vision.is.tohoku.ac.jp}}
\maketitle

\begin{abstract}
Current state-of-the-art methods for image captioning employ region-based features, as they provide object-level information that is essential to describe the content of images; they are usually extracted by an object detector such as Faster R-CNN. However, they have several issues, such as lack of contextual information, the risk of inaccurate detection, and the high computational cost. The first two could be resolved by additionally using grid-based features. However, how to extract and fuse these two types of features is uncharted. This paper proposes a Transformer-only neural architecture, dubbed GRIT (Grid- and Region-based Image captioning Transformer), that effectively utilizes the two visual features to generate better captions. GRIT replaces the CNN-based detector employed in previous methods with a DETR-based one, making it computationally faster. Moreover, its monolithic design consisting only of Transformers enables end-to-end training of the model. This innovative design and the integration of the dual visual features bring about significant performance improvement. The experimental results on several image captioning benchmarks show that GRIT outperforms previous methods in inference accuracy and speed.
\end{abstract}

\section{Introduction}

Image captioning is the task of generating a semantic description of a scene in natural language, given its image. It requires a comprehensive understanding of the scene and its description reflecting the understanding. Therefore, most existing methods solve the task in two corresponding steps; they first extract visual features from the input image and then use them to generate a scene's description. The key to success lies in the problem of how we can extract good features. 

Researchers have considered several approaches to the problem. There are two primary methods, referred to as grid features \cite{xu2015show,rennie2017self,lu2017knowing}
and region features \cite{anderson2018bottom}. Grid features are local image features extracted at the regular grid points, often obtained directly from a higher layer feature map(s) of CNNs/ViTs. Region features are a set of local image features of the regions (i.e., bounding boxes) detected by an object detector.
\begin{figure}[t]
\begin{center}
\includegraphics[width=1.0\linewidth]{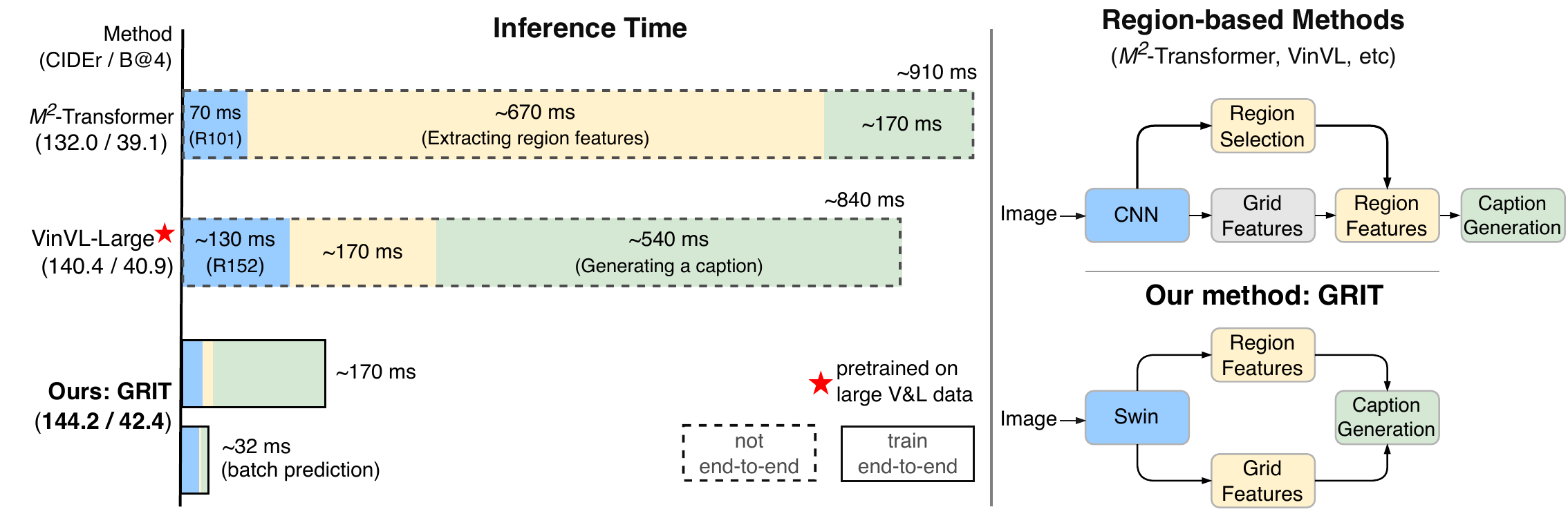}
\end{center}
   \caption{Comparison of GRIT and other region-based methods for image captioning. Left: Running time per image of performing inference with beam size of five and the maximum length of 20 on a V100 GPU. Right: Their architectures
   }
\label{fig:tradeoff}
\end{figure}
The current state-of-the-art methods employ the region features since they encode detected object regions directly. Identifying objects and their relations in an image will be useful to correctly describing the image. However, the region features have several issues. First, they do not convey contextual information such as objects' relation since the regions do not cover the areas between objects. Second, there is a risk of erroneous detection of objects; important objects could be overlooked, etc. Third, computing the region feature is computationally costly, which is especially true when using a high-performance CNN-based detector, such as Faster R-CNN \cite{ren2015faster}. 

The grid features are extracted from the entire image, typically a high-layer feature map of a backbone network. While they do not convey object-level information, they are free from the first two issues with the region features. They may represent contextual information such as objects' relations in images, and they are free from the risk of erroneous object detection. 

In this study, we consider using such region and grid features in an integrated manner, aiming to build a better model for image captioning. The underlying idea is that properly integrating the two types of features will provide a better representation of input images since they are complementary, as explained above. While a few recent studies consider their integration \cite{luo2021dual,xian2022dual}, it is still unclear what the best way is. In this study, we reconsider how to extract each from input images and then consider how to integrate them.

There is yet another issue with the region features, usually obtained by a CNN-based detector. At the last stage of its computation, CNN-based detectors employ non-maximum suppression (NMS) to eliminate redundant bounding boxes. This makes the end-to-end training of the entire model hard, i.e., jointly training the decoder part of the image captioning model and the detector by minimizing a single loss. Recent studies detach the two parts in training; they first train a detector on the object detection task and then train only the decoder part on image captioning. This could be a drag on achieving optimal performance of image captioning. 

To overcome this limitation of CNN-based detectors and also cope with their high-computational cost, we employ the framework of DETR \cite{carion2020end}, which does not need NMS. We choose Deformable DETR \cite{zhu2021deformable}, an improved variant, for its high performance, and also replace a CNN backbone used in the original design with Swin Transformer \cite{liu2021swin} to extract initial features from the input image. We also obtain the grid features from the same Swin Transformer. We input its last layer features into a simple self-attention Transformer and update them to obtain our grid features. This aims to model spatial interaction between the grid features, retrieving contextual information absent in our region features.

The extracted two types of features are fed into the second half of the model, the caption generator. We design it as a lightweight Transformer generating a caption sentence in an autoregressive manner. It is equipped with a unique cross-attention mechanism that computes and applies attention from the two types of visual features to caption sentence words. 

These components form a Transformer-only neural architecture, dubbed GRIT (Grid- and Region-based Image captioning Transformer). 
Our experimental results show that GRIT has established a new state-of-the-art on the standard image captioning benchmark of COCO \cite{lin2014microsoft}. Specifically, in the offline evaluation using the Karpathy test split, GRIT outperforms all the existing methods without vision and language (V\&L) pretraining. It also performs at least on a par with SimVLM$_\mathrm{huge}$ \cite{wang2021simvlm} leveraging V\&L pretraining on 1.8B image-text pairs.

\section{Related Work}

\subsection{Visual Representations for Image Captioning}
Recent image captioning methods typically employ an encoder-decoder architecture. Specifically, given an image, the encoder extracts visual features; the decoder receives the visual features as inputs and generates a sequence of words. Early methods use a CNN to extract a global feature as a holistic representation of the input image \cite{vinyals2015show,karpathy2015deep}. Although it is simple and compact, this holistic representation suffers from information loss and insufficient granularity. To cope with this, several studies \cite{xu2015show,rennie2017self,lu2017knowing} employed more fine-grained grid-based features to represent input images and also used attention mechanisms to utilize the granularity for better caption generation. Later, Anderson et al. \cite{anderson2018bottom} introduced the method of using an object detector, such as Faster R-CNN, to extract object-oriented features, called region features, showing that this leads to performance improvement in many V\&L tasks, including image captioning and visual question answering. Since then, region features have become the de facto choice of visual representation for image captioning. Pointing out the high computational cost of the region features, Jiang et al. \cite{jiang2020defense} showed that the grid features extracted by an object detector perform well on the VQA task. RSTNet \cite{zhang2021rstnet} has recently applied these grid features to image captioning. 

\subsection{Application of Transformer in Vision/Language Tasks}
Transformer has long been a standard neural architecture in natural language processing \cite{vaswani2017attention,devlin2018bert,radford2018improving}, and started to be extended to computer vision tasks. 
Besides ViT \cite{dosovitskiy2020image} for image classification, it was also applied to object detection, leading to DETR \cite{carion2020end}, followed by several variants \cite{zhu2021deformable,fang2021you,song2021vidt}. A recent study \cite{xu2021e2e} applied the framework of DETR to pretraining for various V\&L tasks, where they did not use it to obtain the region features. 

Transformer has been applied to image captioning, where it is used as an encoder for extracting and encoding visual features and a decoder for generating captions. Specifically, Yang et al. \cite{yang2019learning} proposed to use the self-attention mechanism to encode visual features. Li et al. \cite{li2019entangled} used Transformer for obtaining the region features in combination with a semantic encoder that exploits knowledge from an external tagger. Several following studies proposed several variants of Transformer tailored to image captioning, such as Attention on Attention \cite{huang2019attention}, X-Linear Attention \cite{pan2020x}, Memory-augmented Attention \cite{cornia2020meshed}, etc. Transformer is naturally employed also as a caption decoder \cite{herdade2019image,guo2020normalized,luo2021dual,wang2021simvlm}.

\section{Grid- and Region-based Image captioning Transformer}
This section describes the architecture of GRIT (Grid- and Region-based Image captioning Transformer). It consists of two parts, one for extracting the dual visual features from an input image (Sec.~\ref{sec:extraction}) and the other for generating a caption sentence from the extracted features (Sec.~\ref{sec:generation}). 

\begin{figure}[t]
\begin{center}
\includegraphics[width=1.0\linewidth]{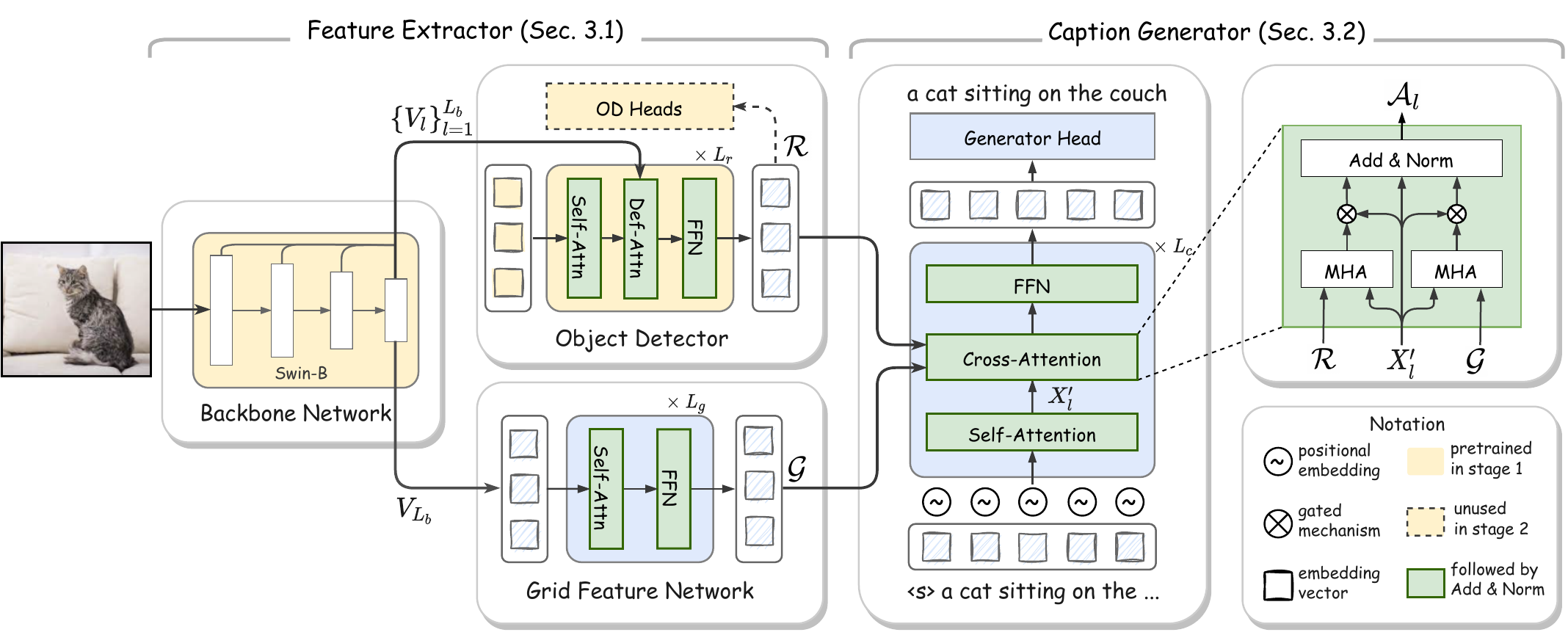}
\end{center}
   \caption{Overview of the architecture of GRIT
   }
\label{fig:overview}
\end{figure}

\subsection{Extracting Visual Features from Images}
\label{sec:extraction}

\subsubsection{Backbone Network for Extracting Initial Features}

A lot of efforts have been made to apply the Transformer architecture to various computer vision tasks since ViT~\cite{dosovitskiy2020image} applied it to image classification. ViT divides an input image into small patches and computes global attention over them. This is not suitable for tasks requiring spatially dense prediction, e.g., object detection since the computational complexity increases quadratically with the image resolution.

Swin Transformer~\cite{liu2021swin} mitigates this issue to a great extent by incorporating operations such as patch reduction and shifted windows that support local attention. It is currently a de facto standard as a backbone network for various computer vision tasks. We employ it to extract initial visual features from the input image in our model. 

We briefly summarize its structure, explaining how we extract features from the input image and send them to the components following the backbone. Given an input image of resolution $H\times W$, Swin Transformer computes and updates feature maps through multiple stages; it uses the patch merging layer after every stage (but the last stage) to downsample feature maps in their spatial dimension by the factor of 2. We apply another patch merging layer to downsample the last layer's feature map. 
We then collect the feature maps from all the stages, obtaining four multi-scale feature maps, i.e., $\{ V_{l}\}_{l=1}^{L_b}$ where ${L_b}=4$, which have the resolution from $H/8 \times W/8$ to $H/64 \times W/64$. These are inputted to the subsequent modules, i.e.,  the object detector and the network for generating grid features.
\subsubsection{Generating Region Features}
As in previous image captioning methods, ours also rely on an object detector to create region features. However, we employ a Transformer-based decoder framework, i.e., DETR \cite{carion2020end} instead of CNN-based detectors, such as Faster R-CNN, which is widely employed by the SOTA image captioning models \cite{anderson2018bottom}. DETR formulates object detection as a direct set prediction problem, which makes the model free of the unideal computation for us, i.e., NMS and RoI alignment. This enables the end-to-end training of the entire model from the input image to the final output, i.e., a generated caption, and also leads to a significant reduction in computational time while maintaining the model’s performance on image captioning compared with the SOTA models. 

Specifically, we employ Deformable DETR \cite{zhu2021deformable}, a variant of DETR. Deformable DETR extracts multi-scale features from an input image with its encoder part, which are fed to the decoder part. We use only the decoder part, to which we input the multi-scale features from the Swin Transformer backbone. This leads to further reduction in computational time. 
We will refer this decoder part as ``object detector’’ in what follows; see Fig.~\ref{fig:overview}.

The object detector receives two inputs: the multi-scale feature maps generated by the backbone, and $N$ learnable object queries $R_0 = \{r_i\}_{i=1}^{N}$, in which $r_i \in \R^d$. Before forwarding them into the object detector, we apply linear transformation to the multi-scale feature maps, mapping them into $d$-dimensional vectors as  $V_l\leftarrow  W_l^r V_l$, where $\{W_l^r\}_{l=1}^{L_b}$ is a learnable projection matrix.

Receiving these two inputs, the object detector updates the object queries through a stack of $L_r$ deformable layers, yielding $R_{L_r}\in \R^{N\times d}$ from the last layer; see \cite{zhu2021deformable} for details. We use $R_{L_r}\in \R^{N\times d}$ as our region features ${\cal R}$. We forward this to the caption generator. 

Although we train it as a part of our entire model, we pretrain our ``object detector'' including the vision backbone on object detection before the training of image captioning.
For the pretraining, we follow the procedure of Deformable DETR; placing a three-layer MLP and a linear layer on its top to predict box coordinates and class category, respectively.
We then minimize a set-based global loss that forces unique predictions via bipartite matching.

Following \cite{anderson2018bottom,zhang2021vinvl}, 
we pretrain the model (i.e., our object detector including the vision backbone) in two steps. 
We first train it on object detection following the training method of Deformable DETR. We then fine-tune it on a joint task of object detection and object attribute prediction, aiming to make it learn fine-grained visual semantics with the following loss:
\begin{equation}
\mathcal{L}_{v}(y,\hat{y}) = \sum_{i=1}^{N}[\underbrace{-{\rm log} \hat{p}_{\hat{\sigma}(i)}(c_i) + \mathbf{1}_{c_i\neq\varnothing}
\mathcal{L}_{box} (b_{i}, \hat{b}_{\hat{\sigma}(i)})}_{\rm object \ detection} \underbrace{-{\rm log} \hat{p}_{\hat{\sigma}(i)}(a_i)}_{\rm attribute\ prediction}],
\end{equation}
where $\hat{p}_{\hat{\sigma}(i)}(a_i)$ and $\hat{p}_{\hat{\sigma}(i)}(c_i)$ are the attribute and class probabilities, $\mathcal{L}_{box}(b_{i}$,$\hat{b}_{\hat{\sigma}(i)})$ is the loss for normalized bounding box regression for object $i$~\cite{zhu2021deformable}.

\subsubsection{Grid Feature Network}

This network receives the last one of the multi-scale feature maps from the Swin Transformer backbone, i.e., $V_{L_b}\in \R^{M \times d_{L_b}}$, where $M = H/64 \times W/64$. As with the input to the object detector, we apply a linear transformation with a learnable matrix $W^g\in \R^{d\times d_{L_b}}$ to $V_{L_b}$, obtaining $G_0= W^g V_{L_b}$
We employ the standard self-attention Transformer having $L_g$ layers. This network updates $V_{L_b}$ through these layers, yielding our grid features $\mathcal{G}$ represented as a $M\times d$ matrix. We intend to extract contextual information hidden in the input image by modeling the spatial interaction between the grid features. 

\subsection{Caption Generation Using Dual Visual Features} 
\label{sec:generation}

\subsubsection{Overall Design of Caption Generator}

The caption generator receives the two types of visual features, the region features ${\cal R} \in \R^{N \times d}$ and the grid features ${\cal G} \in \R^{M \times d}$, as inputs. Apart from this, we employ the basic design employed in previous studies \cite{vaswani2017attention,herdade2019image} that is based on the Transformer architecture. It generates a caption sentence in an autoregressive manner; receiving the sequence of predicted words (rigorously their embeddings) at time $t-1$, it predicts the next word at time $t$. We employ the sinusoidal positional embedding of time step $t$ \cite{vaswani2017attention}; we add it to the word embedding to obtain the input $x^t_0 \in \R^d$ at $t$. 

The caption generator consists of a stack of $L_c$ identical layers. The initial layer receives the sequence of predicted words and the output from the last layer is input to a linear layer whose output dimension equals the vocabulary size to predict the next word.

Each transformer layer has a sub-layer of masked self-attention over the sentence words and a sub-layer(s) of cross-attention between them and the visual features in this order, followed by a feedforward network (FFN) sub-layer.
The masked self-attention sub-layer at the $l$-th layer receives an input sequence $\{{x^{l-1}_i}\}_{i=0}^{t}$ 
at time step $t$, and computes and applies self-attention over the sequence to update the tokens with the attention mask to prevent the interaction from the future words during training.

\begin{figure}[t]
\begin{center}
\includegraphics[width=1.0\linewidth]{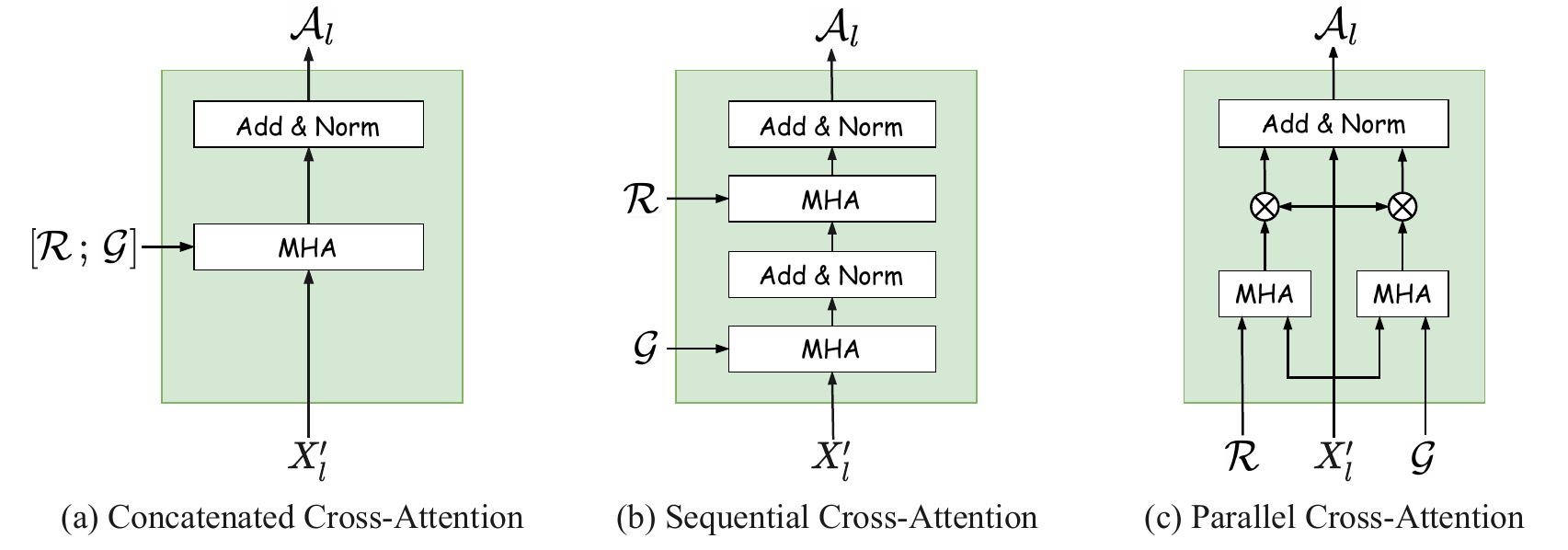}
\end{center}
   \caption{Three designs of cross-attention mechanism to use dual visual features}
\label{fig:cross_attn}
\end{figure}

The cross-attention sub-layer in the layer $l$, located after the self-attention sub-layer, fuses its output with the dual visual features by cross-attention between them, yielding ${\cal A}_l$.
We consider the three design choices shown in Fig.~\ref{fig:cross_attn} and described below. We examine their performance through experiments. 

\subsubsection{Cross-attention between Caption Word and Dual Visual Features}

We show three designs of cross-attention between the word features and the dual visual features (i.e., the region features $\mathcal{R}$ and the grid features $\mathcal{G}$) as below. 

\paragraph{Concatenated Cross-Attention}
The simplest approach is to concatenate the two visual features and use the resultant features as keys and values in the standard multi-head attention sub-layer, where the words serve as queries; see Fig.~\ref{fig:cross_attn}(a).

\paragraph{Sequential Cross-Attention}

Another approach is to perform cross-attention computation separately for the two visual features. The corresponding design is to place two independent multi-head attention sub-layers in a sequential fashion, and uses one for the grid features and the other for the region features (or the opposite combination); see Fig.~\ref{fig:cross_attn}(b). Note that their order could affect the performance. 

\paragraph{Parallel Cross-Attention}
The third approach is to perform multi-head attention computation on the two visual features in parallel. To do so, we use two multi-head attention mechanisms with independent learnable parameters. The detailed design is as follows. Let $X_{l-1}=\{x^{l-1}_i\}$ be the word features inputted to the meta-layer $l$ containing this cross attention sub-layer. As shown in Fig.~\ref{fig:overview}, they are first input to the self-attention sub-layer, converted into $X_l'=\{x_i'\}$ (layer index $l$ omitted for brevity) and then input to this cross attention sub-layer. In this sub-layer, multi-head attention (MHA) is computed with $\{x_i'\}$ as queries and the region features $\mathcal{R}$ as keys and values, yielding attended features $\{a^r_i\}$. The same computation is performed in parallel with the grid features $\mathcal{G}$ as keys and values, yielding $\{a^g_i\}$. Next, we concatenate them with $x_i'$ as $[a^r_i;x_i']$ and $[a^g_i;x_i']$, projecting them back to $d$-dimensional vector using learnable affine projections. Normalizing them with sigmoid into probabilities $\{c_i^r\}$ and $\{c_i^g\}$, respectively, we have 
\begin{align} 
    c_i^g &= \mathrm{sigmoid}(W^g[{a^{g}_{i}}; x^{\prime}_{i}] + b^g), \\
    c_i^r &= \mathrm{sigmoid}(W^r[{a^{r}_{i}}; x^{\prime}_{i}] + b^r).
\end{align}
We then multiply them with $\{a^r_i\}$ and $\{a^g_i\}$, add the resultant vectors to $\{x_i'\}$, and finally feed to layer normalization, obtaining ${\cal A}_l=\{a^{(l)}_i\}$ as follows:
\begin{align} 
    a^{(l)}_{i} &= \mathrm{LN}(c^{g}_i \otimes a^{g}_i + c^{r}_i \otimes a^{r}_i + 
    x^{\prime}_{i}).
    \label{eq:agg}
 \end{align}
\subsubsection{Caption Generator Losses}
Following a standard practice of image captioning studies, we pre-train our model with a cross-entropy loss (XE) and finetune it using the CIDEr-D optimization with self-critical sequence training strategy \cite{rennie2017self}. 
Specifically, the model is first trained to predict the next word $x^*_{t}$ at $t=1..T$, given the ground-truth sentence $x^*_{1:T}$. This is equal to minimize the following XE loss with respect to the model's parameter $\theta$:
\begin{equation}
\mathrm {\cal L}_{XE}(\theta)=-\sum_{t=1}^{T} \log \left(p_{\theta}\left(x_{t}^{*} \mid x^{*}_{0:t-1}\right)\right).
\end{equation}
We then finetune the model with the CIDEr-D optimization, where we use the CIDEr score as the reward and the mean of the rewards as the reward baseline, following \cite{cornia2020meshed}.
The loss for self-critical sequence training is given by
\begin{equation}
    {\cal L}_{RL}(\theta) = -\frac{1}{k}\sum_{i=1}^k (r(\mathbf{w}^i)-b) \log p(\mathbf{w}^i),
\end{equation}
where $\mathbf{w}^i$ is the $i$-th sentence in the beam; $r(\cdot)$ is the reward function; and $b$ is the reward baseline; and $k$ is the number of samples in the batch. 

\section{Experiments}
\subsection{Datasets}
\subsubsection{Object Detection}

As mentioned earlier, we train our object detector (including the backbone) in two steps. In the first step, we train it on object detection using either Visual Genome \cite{krishnavisualgenome} or a combination \cite{zhang2021vinvl} of four datasets: COCO \cite{lin2014microsoft}, Visual Genome, Open Images \cite{OpenImages}, and Object365 \cite{shao2019objects365}, depending on what previous methods we experimentally compare. In the second step, we train the model on object detection plus attribute prediction using Visual Genome. 
Note that following the standard practice, we exclude the duplicated samples appearing in the testing and validation splits of the COCO and {nocaps} \cite{agrawal2019nocaps} datasets to remove data contamination. See the supplementary material for more details.

\subsubsection{Image Captioning}
We conduct our experiments on the COCO dataset, the standard for the research of image captioning  \cite{lin2014microsoft}. The dataset contains 123,287 images, each annotated with five different captions. For offline evaluation, we follow the widely adopted Karpathy split \cite{karpathy}, where 113,287, 5,000, and 5,000 images are used for training, validation, and testing respectively. 

To test our method's effectiveness on other image captioning datasets, we also report the performances on the {nocaps} dataset and the Artemis dataset \cite{achlioptas2021artemis}. See the supplementary material for more details.

\subsection{Implementation Details}
\subsubsection{Evaluation Metrics} 
We employ the standard evaluation protocol for the evaluation of methods. Specifically, we use the full set of captioning metrics: BLEU@N \cite{papineni2002bleu}, METEOR \cite{banerjee2005meteor}, ROUGE-L \cite{lin2004rouge}, CIDEr \cite{vedantam2015cider}, and SPICE \cite{anderson2016spice}.
We will use the abbreviations, B@N, M, R, C, and S, to denote BLEU@N,  METEOR, ROUGE-L, CIDEr, and SPICE, respectively.

\subsubsection{Hyperparameters Settings}
In our model, we set the dimension $d$ of each layer to $512$, the number of heads to eight. We employ dropout with the dropout rate of $0.2$ on the output of {each MHA and FFN sub-layer} following \cite{vaswani2017attention}. 
We set the number of layers as $L_r = 6$ for the object detector, as $L_g=3$ for the grid feature network, and as $L_c=3$ for the caption generator. Following previous studies, we convert all the captions to lower-case, remove punctuation characters, and perform tokenization with the SpaCy toolkit \cite{spacy2}. We build the vocabularies, excluding the words which appear less than five times in the training and validation splits.

\subsection{Training Details}

\subsubsection{First Stage}
In the first stage, we pretrain the object detector with the backbone. We consider several existing region-based methods for comparison, which employ similar pretraining of an object detector but use different datasets. For a fair comparison, we consider two settings. One uses Visual Genome for training, following most previous methods. We train our detector for 150,000 iterations with a batch size of 32. The other (results indicated with $\dagger$ in what follows) uses the four datasets mentioned above,  following \cite{zhang2021vinvl}. We train the detector for 125,000 iterations with a batch size of 256. In both settings, the input image is resized so that the maximum for the shorter side is 800 and for the longer side is 1333. We use Adam optimizer \cite{kingma2015adam} with a learning rate of $10^{-4}$, decreased by 10 at iteration 120,000 and 100,000 in the first and second settings, respectively. We follow \cite{zhu2021deformable} for other training procedures. After this, we finetune the models on object detection plus attribute prediction using Visual Genome for additional five epochs with a learning rate of $10^{-5}$, following \cite{anderson2018bottom,zhang2021vinvl}. The supplementary material presents the details of implementation and experimental results on object detection.

\subsubsection{Second Stage}
We train the entire model for the image captioning task in the second stage. We employ the standard method for word representation, i.e., linear projections of one-hot vectors to vectors of dimension $d$ = 512.  
In this stage, we resize all the input images so that the maximum dimensions for the shorter side and longer side are 384 and 640, respectively. We train models, as explained earlier. Specifically, we train models with the cross-entropy loss ${\cal L}_{XE}$ for ten epochs, in which we warmp up the learning rates for the grid feature network and the caption generator from $10^{-5}$ to $10^{-4}$ in the first epoch, while we fix those for the backbone network and the object detector at $10^{-5}$. Then, we finetune the model based on the CIDEr-D optimization for ten epochs, where we set the fixed learning rate to $5\times 10^{-6}$ for the entire model.
We use the Adam optimizer~\cite{kingma2015adam} with a batch size of $128$. For 
the CIDEr-D optimization, we use beam search with a beam size of 5 and a maximum length of 20. 

\subsection{Performance of Different Configurations}

\begin{table}[t]
    \caption{Results of ablation tests on the COCO test split. All the models are trained with the XE loss and finetuned by the CIDEr optimization.
    }
    \resizebox{1.0\columnwidth}{!}{
    \setlength{\tabcolsep}{2.pt}
    \begin{tabular}[!t]{cccc}
            \begin{tabular}[!t]{l|c|cc}
            \multicolumn{4}{c}{\normalsize \centering (a)}\\
            \toprule
            Factor & Choice & CIDEr & B@4  \\

            \midrule
            (1) \textbf{Backbone Network}       &  ImageNet & 135.5             & 41.5 \\
            \quad - Training data               &  VG       & 142.3             & 41.9 \\
                                                &  4DS   & \textbf{144.2}    & \textbf{42.4} \\
            \midrule
            (2) \textbf{Region features}            & 50     & 141.4       & 41.9 \\
            \quad - Number of vectors               & 100    & 141.8       & 41.5 \\
                    \quad \ \ (trained on VG)                 & 150    & \textbf{142.3}       & \textbf{41.9} \\
            \midrule 
            (3) \textbf{Training strategy}        && \\
            \quad - End-to-end training & Yes  & \textbf{144.2}   & {42.4} \\
             & No  & 139.6   & \textbf{42.7} \\
            \bottomrule 
            \end{tabular}
    & & &
        \begin{tabular}{l|c|cc}
            \multicolumn{4}{c}{\normalsize \centering (b)}\\
            \toprule
            Cross Attention  & Choice & CIDEr & B@4  \\
            \midrule

            (1) \textbf{Concatenated} &  ${\cal G}$             & 142.1 & 41.7 \\
             \quad - Visual features        
                                            & ${\cal R}$             & 142.9 & 41.9 \\
                                            & [${\cal G}$ ; ${\cal R}$]       & 143.1 & 41.9 \\
            \midrule
            (2) \textbf{Sequential} & & \\
            \quad - Sequential order    &  ${\cal G}$ $\rightarrow$ ${\cal R}$  & 144.0 & 42.1 \\
                                &  ${\cal R}$ $\rightarrow$ ${\cal G}$  & 143.6 & 42.1 \\
            \midrule
            (3) \textbf{Parallel} & & \\

            \quad - Gated activation         &  Sigmoid  & \textbf{144.2} & \textbf{42.4} \\
                                          &  Identity & 143.9 & 41.6 \\
        \bottomrule
        \end{tabular}
    \end{tabular}
    \label{tab:ablation}
    
    }
\end{table}

Our method has several design choices. We conduct experiments to examine which configuration is the best. The results are shown in Table \ref{tab:ablation}.
We used an identical configuration unless otherwise noted. Specifically, we use the feature extractor pretrained on the four datasets and parallel cross-attention for fusing the region and grid features.

The first block of Table \ref{tab:ablation}(a) shows the effects of different (pre)training strategies of the visual backbone on image captioning performance. The `ImageNet' column shows the result of the model using a Swin Transformer backbone pretrained on ImageNet21K and the grid features alone; `VG' and `4DS' indicate the models with a detector pretrained on Visual Genome and the four datasets, respectively. They show that using more datasets leads to better performance.

The second block of Table \ref{tab:ablation}(a) shows the effects of the number of object queries, or equivalently region features. The performance increases as they vary as 50, 100, and 150. 
We also confirmed that the performance is saturated for  more region features, while the computational cost and false detection increase. 

The third block shows the effect of the end-to-end training of the entire model. `Yes' indicates the end-to-end training of the entire model and `No' indicates training the model but the vision backbone. The results show that
the end-to-end training considerably improves CIDEr score (from 139.6 to 144.3) with little sacrifice of B@4. This validates our expectation about the effectiveness of the end-to-end training; it arguably helps reduce the domain gap between object detection and image captioning. 

The first block of Table \ref{tab:ablation}(b) shows the performances of the model employing the concatenated cross-attention and its two variants using the grid features alone or the region features alone. They show that the region features alone work better than the grid features alone, and their fusion achieves the highest performance.  

The three blocks of Table \ref{tab:ablation}(b) show the performances of the three cross-attention architectures explained in Sec.~\ref{sec:generation}. The second block shows the two variants of the sequential cross-attention, and the third block shows the two variants of the parallel cross-attention with different gated activation functions, i.e., sigmoid and identity. By identity activation, we mean setting all the values of $c^g_l$ and $c^r_l$ in Eq.(\ref{eq:agg}) to one. These results show that the parallel cross-attention with sigmoid activation function performs the best; the sequential cross-attention in the order ${\cal G} \rightarrow {\cal R}$ attains the second best result. 

\subsection{Results on the COCO Dataset}
We next show complete results on the COCO dataset by the offline and online evaluations. We present example results in the supplementary material.

\begin{table}[!t]
\caption{Offline results evaluated on the COCO Karpathy test split. `V. E. type' indicates the type of visual features;  
`\# VL Data' is the number of image-text pairs used for vision-language pretraining.}
\setlength{\tabcolsep}{5pt}
    \centering
    \scalebox{0.85}{
    \begin{tabular}{l c c c c c c c c c}
    \toprule
        \multirow{2}{*}{Method} &  & {V. E.} & {\# VL} & 
        \multicolumn{6}{c}{Performance Metrics}\\
        \cmidrule(lr){5-10}
         &  & Type & Data &  B@1 & B@4& M & R& C & S\\
        
    \midrule
        {{w/ VL pretraining}}\\
        {UVLP} \cite{zhou2020unified} &  & {${\cal R}$} & {3.0M} & {-} & {39.5} & {29.3} & {-} & {129.3} & {23.2} \\ 
        
        { Oscar$_{\mathrm{base}}$} \cite{li2020oscar} &  &{${\cal R}$} & {6.5M} & {-} & {40.5} & {29.7} & {-} & {137.6} & {22.8} \\
        
        { VinVL$^\dag_{\mathrm{large}}$\cite{zhang2021vinvl}} &  & { ${\cal R}$} & {8.9M} &-& {\textbf{41.0}} & {31.1} & - & {140.9} & {{25.2}} \\
        
        { SimVLM$_{\mathrm{huge}}$ \cite{wang2021simvlm}} &  & {${\cal G}$} & {1.8B} & - & {40.6} & {\textbf{33.7}} & - & {\textbf{143.3}} & {\textbf{25.4}} \\
    
    \midrule
        {w/o VL pretraining}\\
        SAT \cite{vinyals2015show} &  & ${\cal G}$ &- & - &31.9 & 25.5 & 54.3&106.3 & - \\
        SCST \cite{rennie2017self} &  & ${\cal G}$ &- & - &34.2 & 26.7 & 55.7&114.0 & - \\
        
        RSTNet \cite{zhang2021rstnet} &  & ${\cal G}$ & - & 81.8 &  40.1 & 29.8 & 59.5 & 135.6  & 23.0 \\ 
    
        Up-Down \cite{anderson2018bottom} &  & ${\cal R}$ &- &79.8& 36.3 & 27.7 & 56.9&120.1 & 21.4 \\
        RFNet \cite{ke2019reflective} &  & ${\cal R}$ &- &79.1& 36.5 & 27.7 & 57.3&121.9 & 21.2 \\
        GCN-LSTM \cite{yao2018exploring} &  & ${\cal R}$ & -&80.5& 38.2 & 28.5 & 58.3&127.6 &22.0 \\
        LBPF \cite{qin2019look} &  & ${\cal R}$ &- &80.5& 38.3 & 28.5 & 58.4&127.6 & 22.0 \\
        
        SGAE \cite{yang2019auto} &  & ${\cal R}$ &- &80.8& 38.4 & 28.4 & 58.6&127.8 & 22.1 \\
        AoA \cite{huang2019attention} &  & ${\cal R}$ & - &80.2 & 38.9 & 29.2 & 58.8 & 129.8 & 22.4 \\
        NG-SAN \cite{guo2020normalized} &  & ${\cal R}$ & - & - & 39.9 & 29.3 & 59.2 & 132.1 & 23.3 \\
        GET \cite{ji2021improving} &  & ${\cal R}$ & - & 81.5 & 39.5 & 29.3 & 58.9 & 131.6 & 22.8 \\
        
        ORT \cite{herdade2019image} &  & ${\cal R}$ & -& 80.5 & 38.6 & 28.7 & 58.4&128.3 & 22.6 \\
        ETA \cite{li2019entangled} &  & ${\cal R}$ & -& 81.5 & 39.3 & 28.8 & 58.9& 126.6  & 22.6 \\
        ${\cal M}^{2}$ Transformer \cite{cornia2020meshed} &  & ${\cal R}$ & -& 80.8 & 39.1 & 29.2 & 58.6 & 131.2 & 22.6 \\
        X-LAN \cite{pan2020x} &  & ${\cal R}$ & -& 80.8 & 39.5 & 29.5 & 59.2 & 132.0 & 23.4 \\
        TCIC \cite{fan2021tcic} &  & ${\cal R}$ & - & 81.8 & 40.8 & 29.5 & 59.2 & 135.4 & 22.5 \\
        Dual Global \cite{xian2022dual} &  & ${\cal R}$+${\cal G}$ & -& 81.3 & 40.3 & 29.2 & 59.4 & 132.4 & 23.3 \\
        DLCT \cite{luo2021dual} &  & ${\cal R}$+${\cal G}$ &-& 81.4 & 39.8 & 29.5 & 59.1 & 133.8 & 23.0 \\
        \rowcolor{LightCyan}
        GRIT  &  &${\cal R}$+${\cal G}$ & - & 83.5 & 41.9 & 30.5  & 60.5 & 142.2 & {24.2} \\ 
        \rowcolor{LightCyan}
        GRIT$^\dag$ &  &${\cal R}$+${\cal G}$ & - &\textbf{84.2}&\textbf{42.4} &\textbf{30.6} &\textbf{60.7} &\textbf{144.2} & \textbf{24.3} \\ 
    \bottomrule
    \end{tabular}}
    \label{tab:offline_test}
\end{table}

\subsubsection{Offline Evaluation} 
Table \ref{tab:offline_test} shows the performances
of our method and the current state-of-the-art methods on the offline Karpathy test split.
The compared methods are as follows: grid-based methods \cite{vinyals2015show,rennie2017self,yao2017boosting,zhang2021rstnet}, region-based methods \cite{anderson2018bottom,ke2019reflective,yao2018exploring,qin2019look,yang2019auto,huang2019attention,huang2019attention,guo2020normalized,ji2021improving,herdade2019image,li2019entangled,cornia2020meshed,pan2020x,fan2021tcic}, the methods employing both grid and region features \cite{xian2022dual,luo2021dual}, and also the methods relying on large-scale pretraining on vision and language (V\&L) tasks using a large image-text corpus \cite{zhou2020unified,li2020oscar,zhang2021vinvl}, including SimVLM$_\mathrm{huge}$, a model 
pretrained on an extremely large dataset (i.e., 1.8 billion image-caption pairs) \cite{wang2021simvlm}.

For fair comparison with the region-based methods, we report the results of two variants of our model,  one with the object detector pretrained on Visual Genome alone and the other (marked with $^\dagger$) with the object detector pretrained on the four datasets, as explained earlier. 
It is seen from Table \ref{tab:offline_test} that our models, regardless of the datasets used for the detector's pretraining, outperform all the methods that do not use large-scale pretraining of vision and language tasks (i.e., the methods in the second block entitled `w/o VL pretraining'). Moreover, our model with the detector pretrained solely on Visual Genome (i.e., `GRIT') performs better than those relying on large-scale V\&L pretraining but SimVLM$_\mathrm{huge}$. Finally, our model with the pretrained detector on multiple datasets (i.e., `GRIT$^\dagger$') outperforms SimVLM$_\mathrm{huge}$ leveraging large-scale V\&L pretraining in CIDEr score (i.e., 144.2 vs 143.3). 

\subsubsection{Online Evaluation} 
We also evaluate our models (i.e., a single model and an ensemble of six models) on the 40K testing images by submitting their results on the official evaluation server. Table \ref{tab:online_test} shows the results and those of all the published methods on the leaderboard. Table \ref{tab:online_test} presents the metric scores based on five (c5) and 40 reference captions (c40) per image. We can see that our method achieves the best scores for all the metrics. Note that even our single model outperforms all the published methods that use ensembles.

\begin{table}[t]
\caption{Online evaluation results on the COCO image captioning dataset
}
\setlength{\tabcolsep}{3pt}
\centering
\small
\resizebox{1.0\textwidth}{!}{ 
\begin{tabular}{lcllllllllllllll}
\toprule
\multicolumn{1}{l}{\multirow{2}{*}{Method}} & \multirow{2}{*}{Ensemble} & \multicolumn{2}{c}{B-1}   & \multicolumn{2}{c}{B-2}   & \multicolumn{2}{c}{B-3}   & \multicolumn{2}{c}{B-4}   & \multicolumn{2}{c}{M} & \multicolumn{2}{c}{R} & \multicolumn{2}{c}{C} \\
\multicolumn{1}{c}{}& & \multicolumn{1}{c}{c5} & \multicolumn{1}{c}{c40} & \multicolumn{1}{c}{c5} & \multicolumn{1}{c}{c40} & \multicolumn{1}{c}{c5} & \multicolumn{1}{c}{c40} & \multicolumn{1}{c}{c5} & \multicolumn{1}{c}{c40} & \multicolumn{1}{c}{c5} & \multicolumn{1}{c}{c40} & \multicolumn{1}{c}{c5} & \multicolumn{1}{c}{c40} & \multicolumn{1}{c}{c5} & \multicolumn{1}{c}{c40} \\ \hline
{ w/ VL pretraining}\\[0.05cm]
{VinVL$_\mathrm{large}$ \cite{zhang2021vinvl}} & {\xmark} & {81.9} & {96.9} & {66.9} & {92.4} & {52.6} & {84.7} & {40.4} & {74.9} & {30.6} & {40.8} & {60.4} & {76.8} & {134.7}  & {138.7} \\
\midrule
 {w/o VL pretraining}\\
SCST \cite{rennie2017self} & \checkmark & 78.1& 93.7 & 61.9& 86.0 & 47.0& 75.9 & 35.2& 64.5 & 27.0& 35.5 & 56.3& 70.7 & 114.7   & 116.7\\
Up-Down \cite{anderson2018bottom}& \checkmark & 80.2& 95.2 & 64.1& 88.8 & 49.1& 79.4 & 36.9& 68.5 & 27.6& 36.7 & 57.1& 72.4 & 117.9   & 120.5\\
HAN \cite{wang2019hierarchical} &\checkmark & 80.4& 94.5 & 63.8& 87.7 & 48.8   & 78.0& 36.5& 66.8 & 27.4& 36.1 & 57.3& 71.9 & 115.2   & 118.2\\
GCN-LSTM \cite{yao2018exploring} &\checkmark & 80.8& 95.2 & 65.5& 89.3 & 50.8& 80.3 & 38.7& 69.7 & 28.5& 37.6 & 58.5& 73.4 & 125.3   & 126.5\\
SGAE \cite{yang2019auto} & \checkmark &81.0& 95.3 & 65.6& 89.5 & 50.7& 80.4 & 38.5& 69.7 & 28.2& 37.2 & 58.6& 73.6 & 123.8   & 126.5\\
AoA \cite{huang2019attention}  & \checkmark &81.0& 95.0 & 65.8& 89.6 & 51.4& 81.3 & 39.4& 71.2 & 29.1& 38.5 & 58.9& 74.5 & 126.9   & 129.6\\
HIP \cite{Yao2019HierarchyPF} & \xmark &81.6& 95.9 & 66.2& 90.4 & 51.5& 81.6 & 39.3& 71.0 & 28.8& 38.1 & 59.0& 74.1 & 127.9   & 130.2\\
${\cal M}^{2}$Trans. \cite{cornia2020meshed}  &\checkmark & 81.6& 96.0 & 66.4& 90.8 & 51.8& 82.7 & 39.7& 72.8 & 29.4& 39.0 & 59.2& 74.8 & 129.3   & 132.1\\
X-LAN \cite{pan2020x} &\checkmark & 81.9& 95.7 & 66.9& 90.5 & 52.4& 82.5 & 40.3& 72.4 & 29.6& 39.2 & 59.5& 75.0 & 131.1   & 133.5   \\
Dual Global \cite{xian2022dual} &\xmark & {80.8}  & {95.1}   & {65.6}  & {81.3} & 51.1 & 81.3 &  {39.1}  & {71.2}   & {28.9}  & 38.4 & 58.9 & 74.4 & 126.3 & 129.2 \\
DLCT \cite{luo2021dual} &\checkmark & {82.4}  & {96.6}   & {67.4}  & {91.7}   & {52.8}  & {83.8}   & {40.6}  & {74.0}   & {29.8}  & {39.6}   & {59.8}  & {75.3}   & {133.3} & {135.4}  \\ 
\rowcolor{LightCyan}
GRIT$^\dag$ & \xmark & {83.7} & {97.4}   & {68.5}  & {92.8}   & {53.9}  & {85.3}   & {41.5}  & {75.6}   & {30.3}  & {40.2}   & {60.2}  & {75.9}   & {138.3} & {141.8}  \\ 
\rowcolor{LightCyan}
GRIT$^\dag$ & \checkmark & \textbf{84.1} & \textbf{97.6}   & \textbf{69.4}  & \textbf{93.5}   & \textbf{54.9}  & \textbf{86.3}   & \textbf{42.5}  & \textbf{76.8}   & \textbf{30.9}  & \textbf{41.0}   & \textbf{61.2}  & \textbf{77.1}   & \textbf{141.3} & \textbf{143.8}  \\ 
\bottomrule
\label{tab:online_test}
\end{tabular}
}
\end{table}

\subsection{Results on the ArtEmis and {nocaps} Datasets}
As explained above, we evaluate our method on the ArtEmis and nocaps datasets. For nocaps, we evaluate zero-shot inference performance, i.e., the performance of the model trained on COCO. For ArtEmis, we train the model in the same way as COCO except for the number of training epochs, precisely, five epochs each for the training with the XE loss and that with the CIDEr-D optimization.

Table \ref{tab:other}(a) shows the results of our method on the test split of ArtEmis \cite{achlioptas2021artemis}. It also show the results of existing methods { reported in \cite{achlioptas2021artemis}}, which are grid-based \cite{mathews2016senticap,vinyals2015show}, region-based \cite{cornia2020meshed}, and a nearest neighbor method using a holistic vector to encode images (denoted as ${\cal H}$).  Our method outperforms all these methods by a large margin.

Table \ref{tab:other} shows the results on the nocaps dataset, including the baseline methods reported in \cite{agrawal2019nocaps,cornia2020meshed}. All the models are trained on the training split of the COCO datasets and tested on the validation split of nocaps, which consists of images with novel objects and captions with unseen vocabularies.
Our method surpasses all the other methods including region-based methods \cite{lu2018neural,anderson2018bottom,cornia2020meshed} in both in-domain and out-of-domain images. See the supplementary material for the full results.


\begin{table}[t]
    \caption{Performance on the ArtEmis and {nocaps} datasets}
    \vskip -0.05in
    \resizebox{1.0\columnwidth}{!}{
    \setlength{\tabcolsep}{1.pt}
\begin{tabular}[!t]{cccc}
    \begin{tabular}{l c c c c c c c}
        \multicolumn{8}{c}{a) Performance on the ArtEmis test split}\\
        \toprule
        \multirow{2}{*}{Method} & 
        {V. E.} & 
        \multicolumn{6}{c}{Performance Metrics}\\
        \cmidrule(lr){3-8}
         & Type &  B@1 & B@2 & B@3 & B@4 & M & R \\
        \midrule
        NN \cite{achlioptas2021artemis} & ${\cal H}$ & 36.4 & 13.9 & 5.4 & 2.2 & 10.2 & 21.0 \\
        ANP \cite{achlioptas2021artemis} & ${\cal G}$ & 39.6 & 13.4 & 4.2 & 1.4 & 8.8 & 20.2 \\
        SAT \cite{achlioptas2021artemis} & ${\cal G}$ & 53.6 & 29.0 & 15.5 & 8.7 & 14.2 & 29.7  \\
        ${\cal M}^{2}$Trans. \cite{achlioptas2021artemis} & ${\cal R}$ & 50.7  & 28.2  & 15.9 & 9.5 & 13.7 & 28.0 \\
        \rowcolor{LightCyan}
        GRIT$^\dag$ &${\cal R}$+${\cal G}$ &\textbf{70.1}&\textbf{40.1} &\textbf{20.9} &\textbf{11.3} &\textbf{16.8} & \textbf{33.3} \\ 
    \bottomrule
    \end{tabular}
    & & &
    \begin{tabular}{l c c c c c c c}
        \multicolumn{8}{c}{b) Performance on the {nocaps} validation split}\\
        \toprule
        \multirow{2}{*}{Method}  &V.E& \multicolumn{2}{c}{In-Domain} & \multicolumn{2}{c}{Out-Domain} & \multicolumn{2}{c}{Overall}  \\
        \cmidrule(lr){3-4} \cmidrule(lr){5-6} \cmidrule(lr){7-8}
         & Type &  C & S & C & S & C & S \\
        \midrule
        NBT \cite{agrawal2019nocaps} & ${\cal R}$ & 62.7 & 10.1 & 54.0 & 8.6 & 53.9 & 9.2 \\
        Up-down \cite{agrawal2019nocaps} & ${\cal R}$ & 78.1 & 11.6 & 31.3 & 8.3 & 55.3 & 10.1 \\
        Trans. \cite{cornia2020meshed} & ${\cal R}$ & 78.0 & 11.0 & 29.7 & 7.8 & 54.7 & 9.8  \\
        ${\cal M}^{2}$Trans. \cite{cornia2020meshed} & ${\cal R}$ & 85.7  & 12.1  & 38.9 & 8.9 & 64.5 & 11.1 \\
        \rowcolor{LightCyan}
        GRIT$^\dag$ &${\cal R}$+${\cal G}$  &\textbf{105.9}&\textbf{13.6} &\textbf{72.6} &\textbf{11.1} &\textbf{90.2} & \textbf{12.8} \\ 
    \bottomrule
    \end{tabular}
\end{tabular}

    \label{tab:other}
    
    }
\end{table}

\subsection{Computational Efficiency} 
We measured the inference time of GRIT and two representative region-based methods, VinVL \cite{zhang2021vinvl} and ${\cal M}^2$ Transformer \cite{cornia2020meshed}. It is the computational time per image from image input to caption generation. Specifically, we measured the time to generate a caption of length 20 with a beam size of five on a V100 GPU. The input image resolution was set to $800 \times 1333$ for VinVL and ${\cal M}^2$ Transformer as reported in \cite{anderson2018bottom,zhang2021vinvl}. We set it to $384\times640$ for GRIT since it already achieves higher accuracy. Figure ~\ref{fig:tradeoff} shows the breakdown of the inference time for the three methods. GRIT reduces the time for feature extraction by a factor of 10 compared with the others. 
Similar to ${\cal M}^2$ Transformer, GRIT has a lightweight caption generator and thus spends much 
less time than VinVL for generating a caption after receiving the visual features.
GRIT can run with minibatch size up to $64$ on a single V100 GPU, while others cannot afford large minibatch. With minibatch size $\geq32$, the per-image inference time decreases to about 32ms. More details are given in the supplementary material.

\section{Summary and Conclusion}
In this paper, we have proposed a Transformer-based architecture for image captioning named GRIT. It integrates the region features and the grid features extracted from an input image to extract richer visual information from input images. Previous SOTA methods employ a CNN-based detector to extract region features, which prevents the end-to-end training of the entire model and makes to high computational costs. Using the Swin Transformer for a backbone extracting the initial visual feature, GRIT resolves these two issues by employing a DETR-based detector. Furthermore, GRIT obtains grid features by updating the feature from the same backbone using a self-attention Transformer, aiming to extract richer context information complementing the region feature. These two features are fed to the caption generator equipped with a unique cross-attention mechanism, which computes and applies attention from the dual features on the generated caption sentence. The integration of all these components led to significant performance improvement. The experimental results validated our approach, showing that GRIT outperforms all published methods by a large margin in inference accuracy and speed. 

\subsubsection{Acknowledgments} This work was supported by JST [Moonshot Research and Development], Grant Number [JPMJMS2032] and by JSPS KAKENHI Grant Number 20H05952 and 19H01110.

\appendix
\section{Additional Details for Object Detection}
\subsection{Object Detection Datasets}
When pretraining our model on the four datasets (i.e., Visual Genome (VG), COCO, OpenImages, and Objects365), we follow \cite{zhang2021vinvl} to build a unified training corpus with the statistics shown in Table \ref{tab:stat} except that we do not use the annotations from COCO stuff \cite{caesar2018cvpr}. The resultant corpus has images with 1848 categories.
\begin{table}[h]
    \centering
    \setlength{\tabcolsep}{3.pt}
    \caption{Statistics of the pretraining datasets for object detection.}
    \begin{tabular}{l c  c c c  c}
        \toprule
        Source & &  VG & COCO & Objects365 & OpenImages \\
        \midrule
        Images  & &  97k & 111k & 609k & 1.67M \\
        Categories & &  1594 & 80 & 365 & 500 \\
        Sampling & & $\times 8$ & $\times 8$ & $\times 2$ & $\times 1$ \\
        \bottomrule
    \end{tabular}
    \label{tab:stat}
\end{table}
\subsection{Implementation Details}
For the object detector, we set the number of queries $N=150$, the number of sampling points equal to 4, and the hidden dimension $d=512$. The backbone network weights are intialized by the weights of Swin-Base ($384\times384$) pretrained on ImageNet21K \cite{liu2021swin}. Following \cite{zhu2021deformable}, the loss for normalized bounding box regression for object $i$, $\mathcal{L}_{box}(b_{i}$,$\hat{b}_{\hat{\sigma}(i)})$ is computed as the weighted summation of a box distance ${\cal L}_{l_1}$ and a GIoU loss ${\cal L}_{iou}$:
\begin{align} 
    {\cal L}_{l_1}(b_{i},\hat{b}_{\hat{\sigma}(i)}) &= ||b_i - \hat{b}_{\hat{\sigma}(i)}||_1, \\
    {\cal L}_{iou}(b_{i},\hat{b}_{\hat{\sigma}(i)}) &= 1 -  \big( \frac{|b_{i} \cap  \hat{b}_{\sigma(i)}|}{|b_{i} \cup  \hat{b}_{\sigma(i)}|} - \frac{|{\sf B}(b_{i}, \hat{b}_{\sigma(i)}) \symbol{92} b_{i} \cup  \hat{b}_{\sigma(i)}| }{|{\sf B}(b_{i}, \hat{b}_{\sigma(i)})|} \big), \\
    \mathcal{L}_{box}(b_{i},\hat{b}_{\hat{\sigma}(i)}) &= \alpha_{l_1}{\cal L}_{l_1}(b_{i},\hat{b}_{\hat{\sigma}(i)}) + \alpha_{iou}{\cal L}_{iou}(b_{i},\hat{b}_{\hat{\sigma}(i)}),
\end{align}
where $\alpha_{l_1} = 5$, $\alpha_{iou} = 2$, and ${\sf B}$ outputs the largest box covering $b_{i}$ and $\hat{b}_{\hat{\sigma}(i)}$. We also employ two training strategies, i.e., iterative bounding box refinement and auxiliary losses; see \cite{zhu2021deformable} and our configuration files for details.

\begin{table}[h]
    \centering
    \caption{Performance of object detection on the COCO and Visual Genome datasets. `4DS' denotes the four object detection datasets.}
    \setlength{\tabcolsep}{3.pt}
    \begin{tabular}{l c  c c c  c}
        \toprule
        Model & &  Training Data & mAP (COCO) & mAP$^{50}$ (VG)  \\
        \midrule
        BUTD \cite{anderson2018bottom} & &  VG & - & 10.2 \\
        VinVL \cite{zhang2021vinvl} & &  4DS & 50.5 & 13.8 \\
        \rowcolor{LightCyan}
        GRIT & & VG  & 33.6 & 14.2 \\
        \rowcolor{LightCyan}
        GRIT$^\dag$ & & 4DS & 50.8 & 15.1 \\
        \bottomrule
    \end{tabular}
    \label{tab:od_results}
\end{table}

\subsection{Object Detection Results}
Table \ref{tab:od_results} shows the performance on the COCO validation split and the Visual Genome test split of our object detector compared with VinVL and BUTD \cite{anderson2018bottom}. It is seen that the object detector of GRIT attains comparable or higher performance on the two datasets as compared with BUTD and VinVL when pretrained on the similar datasets.

\section{Additional Details for Image Captioning}
\subsubsection{Class Token} We prepend a class token embedding $g_{\langle \texttt{cls}\rangle} \in \R^d$ to $G_0$ before forwarding them to the grid feature network. We use this class token embedding to predict the emotion category of the input image when training an emotion-grounded model on the ArtEmis dataset; see Sec. \ref{sec:artemis}.

\subsubsection{Boundary Tokens} Following previous studies, we prepend a special token $\langle \texttt{sos}\rangle$ to the beginning of captions, and append another special token $\langle \texttt{eos}\rangle$ to the end of captions during training. During inference, we start the generation by setting the first token to $\langle \texttt{sos}\rangle$.
\begin{table}[h]
    \centering
    \setlength{\tabcolsep}{3.pt}
    \caption{Breakdown of SPICE F-scores over various sub-categories and the CLIP scores.}
    \begin{tabular}{l c c c c c c c c} 
        \toprule
        Method &  SPICE & Object & Attr. & Relation & Color & Count & Size & CLIP \\
        \midrule
        Up-Down \cite{anderson2018bottom} & 21.4 & 39.1 & 10.0 & 6.5 & 11.4 & 18.4 & 3.2 & - \\
        Transformer \cite{cornia2020meshed} & 21.1 & 38.6 & 9.6 & 6.3 & 9.2 & 17.5 & 2.0 & - \\
        ${\cal M}^2$ Trans. \cite{cornia2020meshed} & 22.6 & 40.0 & 11.6 & 6.9 & 12.9 & 20.4 & 3.5 & 73.4 \\
        \rowcolor{LightCyan}
        GRIT$^\dag$ & 24.3 & 42.7 & 13.5 & 7.7 & 14.7 & 29.3 & 4.5 & 77.2 \\
        \bottomrule
    \end{tabular}
    
    \label{tab:spice}
\end{table}

\subsection{Image Captioning on the COCO dataset}
\subsubsection{SPICE Sub-category and CLIPscore Metrics} Table \ref{tab:spice} reports a breakdown of
SPICE F-scores over various sub-categories on the ``Karpathy'' test split, in comparison with the region-based methods: Up-Down \cite{anderson2018bottom}, vanilla Transformer \cite{cornia2020meshed}, and ${\cal M}^2$ Transformer \cite{cornia2020meshed}. These scores give a quantitative assessment of performance on different aspects when describing the content of images. 
As seen in Table \ref{tab:spice}, our method attains better scores over all sub-categories, showing significant improvement on identifying and counting objects, attributes, and relationships between objects.
The table also reports the CLIP scores \cite{hessel2021clipscore} of the two methods, showing consistent improvement of our method over the compared method.
\subsection{Image Captioning on the ArtEmis dataset} \label{sec:artemis}
\subsubsection{ArtEmis Dataset} This dataset consists of 80,031 unique images divided into the training, validation, and test splits with the ratios of 85\%, 5\%, and 10\%, respectively. Each caption of a given image is annotated with an emotion label. In total, there are 454,684 captions along with 8 unique emotion categories; see \cite{achlioptas2021artemis} for details.

\subsubsection{Emotion Grounded Model} Following \cite{achlioptas2021artemis}, we also trained an emotion grounded model, which predicts the emotion associated with the caption. Specifically, we mapped the updated class embedding $g_{\langle \texttt{cls}\rangle}$ into an $8$-dimensional vector using a linear projection. During training, we minimized the summation of the two losses, i.e., emotion prediction and caption generation.

\subsubsection{Full Results} Table \ref{tab:artemis_res} shows the full results of different models on the test split of the Artemis dataset including the emotion grounded models. It is noted that the ground truth emotion labels are not provided during inference. 
\begin{table}[h]
    \centering
    \caption{Performance on the ArtEmis test split.}
    \setlength{\tabcolsep}{2.pt}
    \resizebox{0.8\columnwidth}{!}{
    \begin{tabular}{l c c c c c c c c}
        \toprule
        \multirow{2}{*}{Method} & Emotion &
        {V. E.} & 
        \multicolumn{6}{c}{Performance Metrics}\\
        \cmidrule(lr){4-9}
         &Grounded & Type &  B@1 & B@2 & B@3 & B@4 & M & R \\
        \midrule
        NN \cite{achlioptas2021artemis} & No & ${\cal H}$ & 36.4 & 13.9 & 5.4 & 2.2 & 10.2 & 21.0 \\
        ANP \cite{achlioptas2021artemis} & No & ${\cal G}$ & 39.6 & 13.4 & 4.2 & 1.4 & 8.8 & 20.2 \\
        ${\cal M}^{2}$Trans. \cite{achlioptas2021artemis}  &Yes & ${\cal R}$ & 51.1  & 28.2  & 15.4 & 9.0 & 13.7 & 28.6 \\
        ${\cal M}^{2}$Trans. \cite{achlioptas2021artemis} & No & ${\cal R}$ & 50.7  & 28.2  & 15.9 & 9.5 & 14.0 & 28.0 \\
        SAT \cite{achlioptas2021artemis} & Yes & ${\cal G}$ & 52.0 & 28.0 & 14.6 & 7.9 & 13.4 & 29.4  \\
        SAT \cite{achlioptas2021artemis} & No & ${\cal G}$ & 53.6 & 29.0 & 15.5 & 8.7 & 14.2 & 29.7  \\

        \rowcolor{LightCyan}
        GRIT$^\dag$ & Yes &${\cal R}$+${\cal G}$ &{69.3}&{39.4} 
         &{19.2}&{11.1} & {16.5} & {33.0} \\         
        \rowcolor{LightCyan}
        GRIT$^\dag$ & No &${\cal R}$+${\cal G}$  &\textbf{70.1}&\textbf{40.1} &\textbf{20.9} &\textbf{11.3} &\textbf{16.8} & \textbf{33.3} \\ 
        
    \bottomrule
    \end{tabular}
    }
    \label{tab:artemis_res}
\end{table}

\subsection{Image Captioning on the nocaps Dataset}
\subsubsection{Full results} 
We report the full results on the validation split of the nocaps dataset for different domains, i.e., in-domain, near-domain, and out-of-domain, in Table \ref{tab:nocaps}.

\begin{table}[h]
    \centering
    \caption{Performance on the {nocaps} validation split.}
    \setlength{\tabcolsep}{2.pt}
    \resizebox{0.8\columnwidth}{!}{
    \begin{tabular}{l c c c c c c c c c}
        \toprule
        \multirow{2}{*}{Method}  &V.E& \multicolumn{2}{c}{in-domain} & \multicolumn{2}{c}{near-domain} & \multicolumn{2}{c}{\small {out-domain}} & \multicolumn{2}{c}{Overall}  \\
        \cmidrule(lr){3-4} \cmidrule(lr){5-6} \cmidrule(lr){7-8} \cmidrule(lr){9-10}
         & Type &  C & S & C & S & C & S & C & S \\
        \midrule
        NBT \cite{agrawal2019nocaps} & ${\cal R}$ & 62.7 & 10.1 & 51.9 & 9.2 & 54.0 & 8.6 & 53.9 & 9.2 \\
        Up-down \cite{agrawal2019nocaps} & ${\cal R}$ & 78.1 & 11.6 & 57.7 & 10.3 & 31.3 & 8.3 & 55.3 & 10.1 \\
        Trans. \cite{cornia2020meshed} & ${\cal R}$ & 78.0 & 11.0 & - & - & 29.7 & 7.8 & 54.7 & 9.8  \\
        ${\cal M}^{2}$Trans. \cite{cornia2020meshed} & ${\cal R}$ & 85.7  & 12.1 &- & - & 38.9 & 8.9 & 64.5 & 11.1 \\
        \rowcolor{LightCyan}
        GRIT$^\dag$ &${\cal R}$+${\cal G}$  &\textbf{105.9}&\textbf{13.6} & \textbf{92.16} & \textbf{13.05} &\textbf{72.6} &\textbf{11.1} &\textbf{90.2} & \textbf{12.8} \\ 
    \bottomrule
    \end{tabular}

    \label{tab:nocaps}
    }
\end{table}

\begin{figure}[h]
\begin{center}
\includegraphics[width=.9\linewidth]{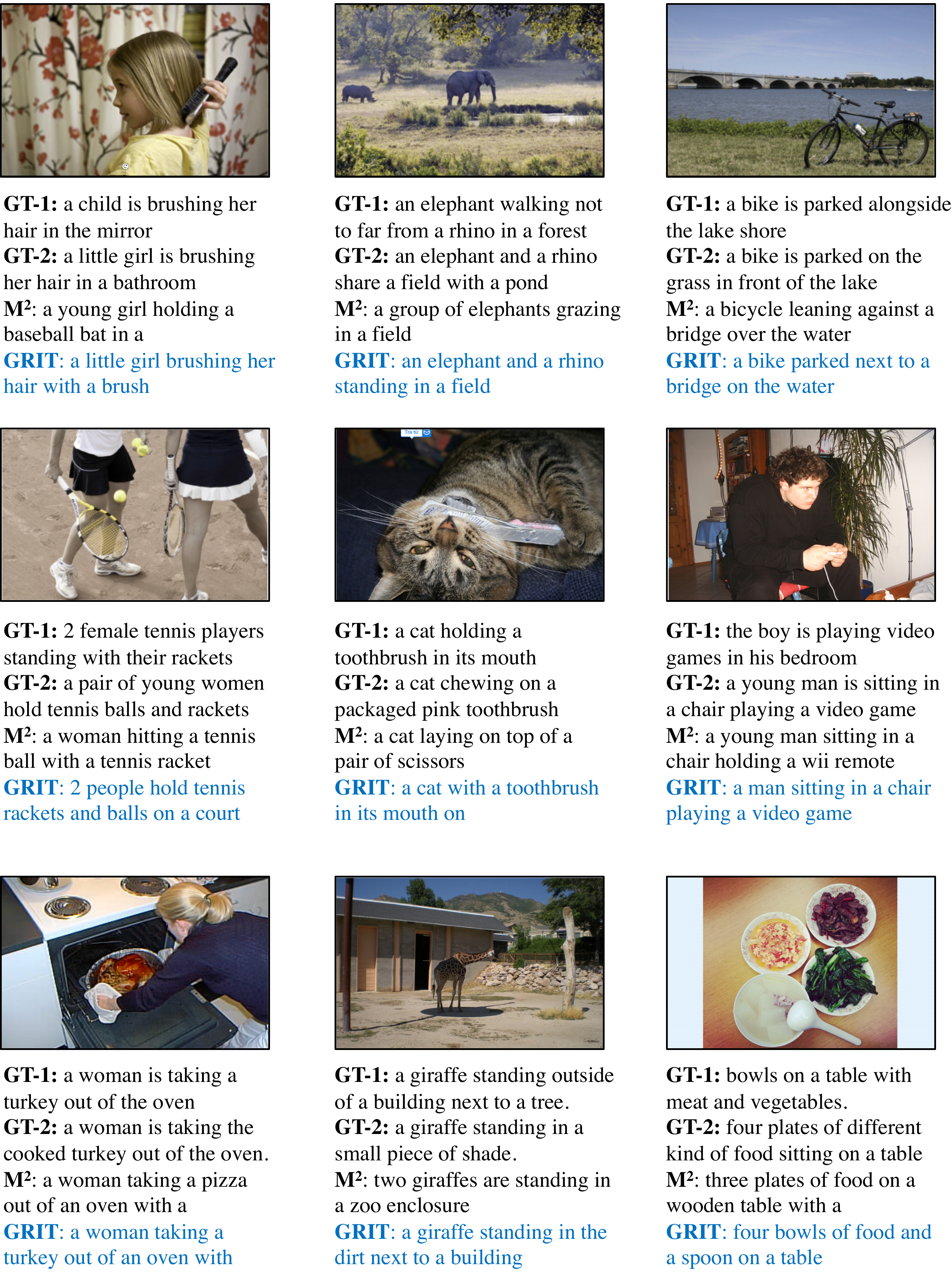}
\end{center}
   \caption{Qualitative examples from our method (GRIT) and a region-based method (${\cal M}^2$ Transformer) on the COCO test images. Zoom in for  better view.}
\label{fig:res1}
\end{figure}

\begin{figure}[h]
\begin{center}
\includegraphics[width=.9\linewidth]{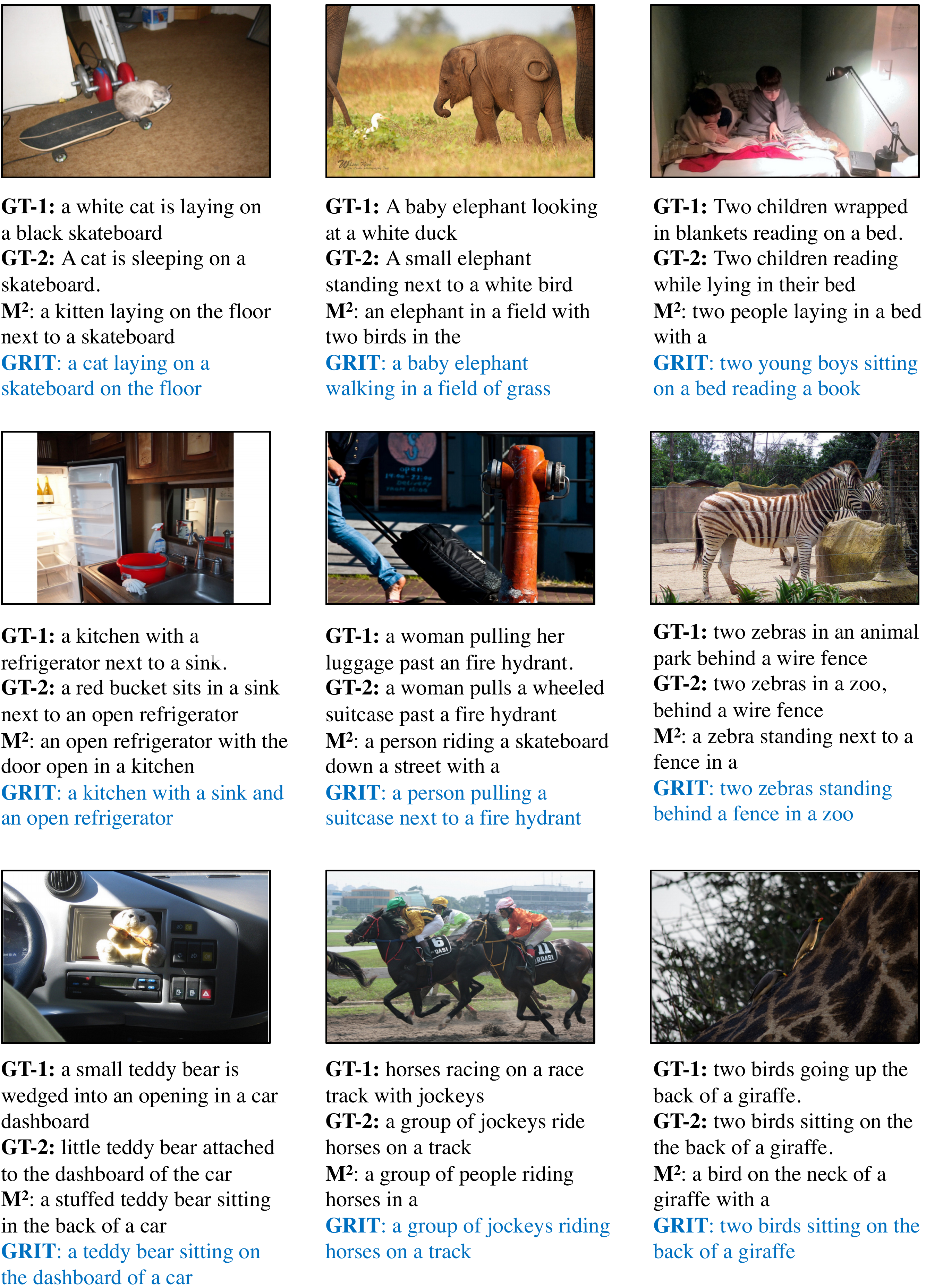}
\end{center}
   \caption{Qualitative examples from our method (GRIT) and a region-based method (${\cal M}^2$ Transformer) on the COCO test images. Zoom in for  better view.}
\label{fig:res2}
\end{figure}
\begin{figure}[h]
\begin{center}
\includegraphics[width=.9\linewidth]{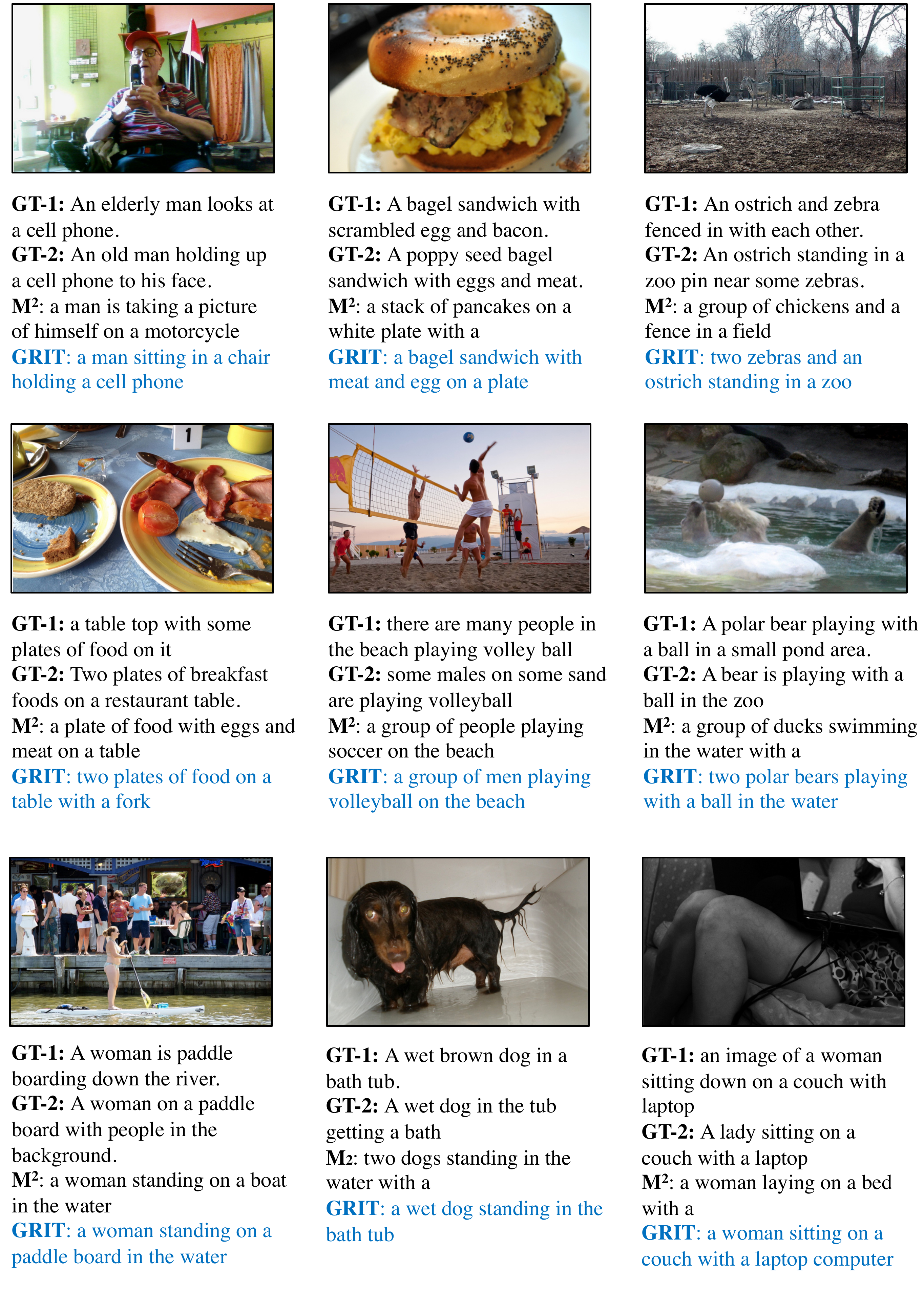}
\end{center}
   \caption{Qualitative examples from our method (GRIT) and a region-based method (${\cal M}^2$ Transformer) on the COCO test images. Zoom in for  better view.}
\label{fig:res3}
\end{figure}
\begin{figure}[h]
\begin{center}
\includegraphics[width=.75\linewidth]{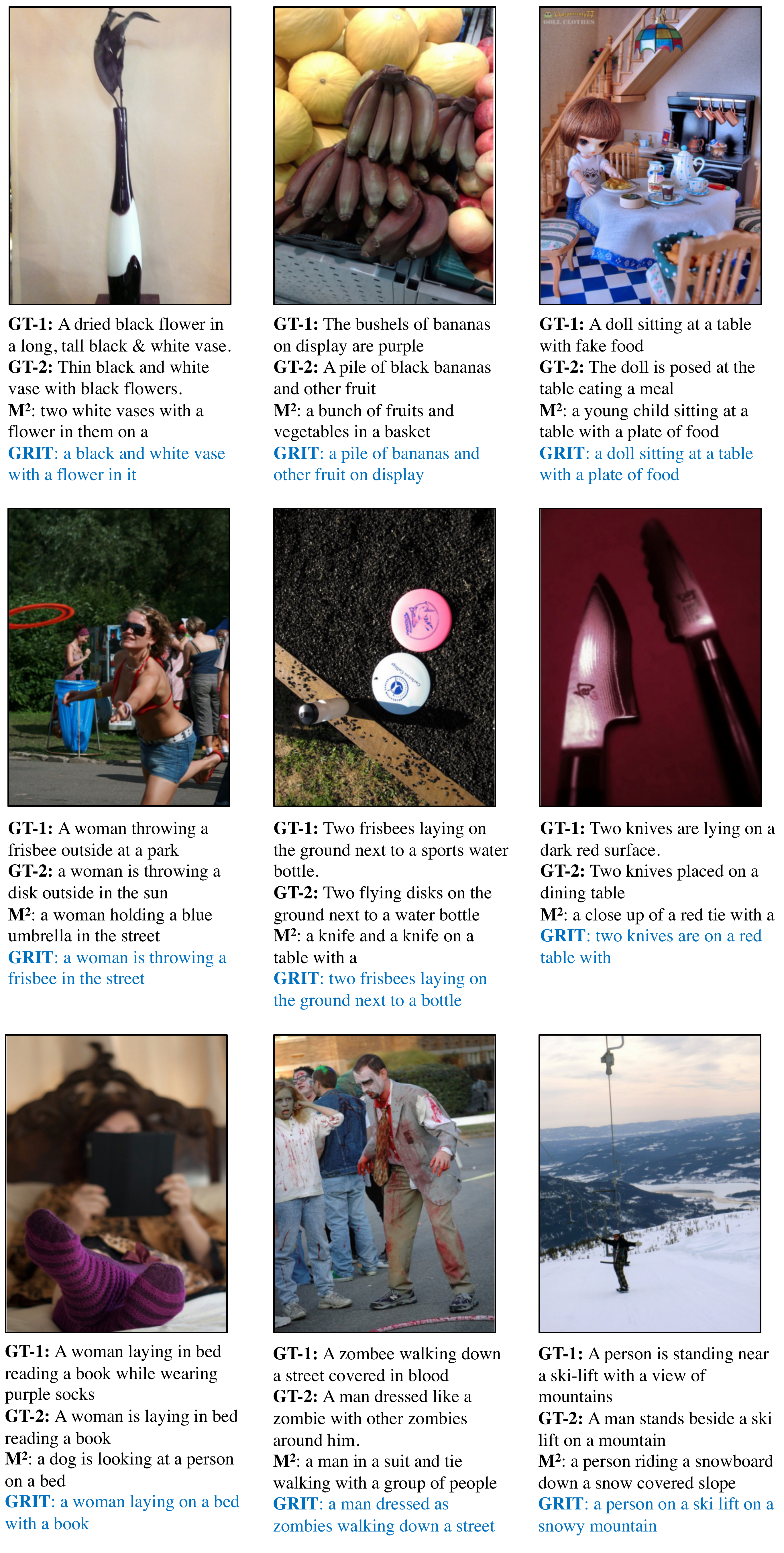}
\end{center}
   \caption{Qualitative examples from our method (GRIT) and a region-based method (${\cal M}^2$ Transformer) on the COCO test images. Zoom in for  better view.}
\label{fig:res4}
\end{figure}

\subsection{Computational Efficiency} 
We measured the inference time of GRIT and two representative region-based methods, VinVL \cite{zhang2021vinvl} and ${\cal M}^2$ Transformer \cite{cornia2020meshed}, on the same machine having a Tesla V100-SXM2 of 16GB memory with CUDA version 10.0 and Driver version 410.104. It has Intel(R) Xeon(R) Gold 6148 CPU. The comparison was conducted following \cite{jiang2020defense,kim2021vilt}. Specifically, we excluded the time of preprocessing the image and loading it to the GPU device. Also, the images are rescaled to the resolutions such that all the compared methods achieve its highest performance for image captioning.
For the compared methods, we used the official implementations of ${\cal M}^2$ Transformer\footnote{\url{https://github.com/aimagelab/meshed-memory-transformer}} and VinVL\footnote{\url{https://github.com/pzzhang/VinVL}}.

Regarding feature extraction, we extracted the region features from Faster R-CNN using the original implementation\footnote{\url{https://github.com/peteanderson80/bottom-up-attention}} used by ${\cal M}^2$ Transformer and another implementation\footnote{\url{https://github.com/microsoft/scene\_graph\_benchmark}} used by VinVL.
It is seen that VinVL and ${\cal M}^2$ Transformer spend considerable time on feature extraction due to the forward pass through the CNN backbone with high resolution inputs and the computationally expensive regional operations. It is also noted that VinVL introduced class-agnostic NMS operations, which reduce a great amount of time consumed by class-aware NMS operations in the standard Faster R-CNN. On the other hand, we employ a Deformable DETR-based detector to extract region features without using all such operations. Table \ref{tab:extraction} shows the comparison on feature extraction.

\begin{table}[h]
    \centering
    \setlength{\tabcolsep}{3.pt}
    \caption{The inference time on feature extraction of different methods.}
    \begin{tabular}{l c c c c c c c r} 
        \toprule
        Method &  Backbone & Detector & Regional Operations & Inference Time \\
        \midrule
        VinVL$_\mathrm{large}$\cite{zhang2021vinvl} & ResNeXt-152 & Faster R-CNN & Class-Agnostic NMS & 304 ms \\
        & & & RoI Align, etc &  \\
        ${\cal M}^2$ Trans. \cite{cornia2020meshed} & ResNet-101 & Faster R-CNN & Class-Aware NMS & 736 ms \\
        & & & RoI Align, etc &  \\
        \rowcolor{LightCyan}
        GRIT & Swin-Base & DETR-based & - & 31 ms \\
        \bottomrule
    \end{tabular}
    \label{tab:extraction}
\end{table}

Regarding caption generation, all the methods use beam search as the decoding strategy, with beam size of 5 and the maximum caption length of 20. 
Both ${\cal M}^2$ Transformer and GRIT employ a lightweight caption generator (caption decoder) having only 3 transformer layers with hidden dimension of 512 while VinVL$_\mathrm{large}$ has 24 transformer layers with hidden dimension of 1024; see Table \ref{tab:generator}. Thus, with the visual features as inputs, ${\cal M}^2$ Transformer and GRIT spend less inference time generating words than VinVL$_\mathrm{large}$ in the autoregressive manner.

\begin{table}[h]
    \centering
    \setlength{\tabcolsep}{3.pt}
    \caption{The inference time on caption generation of different methods.}
    \begin{tabular}{l c c c c c c c} 
        \toprule
        Method &  No. of Layers & Hidden Dim. & Inference Time \\
        \midrule
        VinVL$_\mathrm{large}$\cite{zhang2021vinvl} & 24 & 1024 & 542 ms \\
        ${\cal M}^2$ Transformer \cite{cornia2020meshed} & 3 & 512 & 174 ms \\
        \rowcolor{LightCyan}
        GRIT & 3 & 512 & 138 ms \\
        \bottomrule
    \end{tabular}
    \label{tab:generator}
\end{table}

\subsection{Qualitative Examples} 
Figure \ref{fig:res1}, \ref{fig:res2}, \ref{fig:res3}, and \ref{fig:res4} show some examples of the captions generated by our proposed method (GRIT) and another region-based method (${\cal M}^2$ Transformer) given the same input images from the COCO test split. 
It is observed that the generated captions from GRIT are qualitatively better than those generated by the baseline method in terms of detecting and counting objects as well as describing their relationships in the given images. The inaccuracy of the captions generated by the baseline method might be due to the drawbacks of the region features extracted by a frozen pretrained object detector which produces wrong detection and lacks of contextual information.

\bibliographystyle{splncs04}
\bibliography{egbib}
\end{document}